\documentclass[12pt]{article}

\usepackage{times}
\usepackage{url}
\usepackage{graphicx}
\usepackage{amsmath, amssymb}
\usepackage{cite}
\usepackage{hyperref}
\usepackage{geometry}
\usepackage{authblk}
\usepackage{adjustbox}
\title{Placenta Accreta Spectrum Detection Using an MRI-based Hybrid CNN-Transformer Model}

\author[1*]{Sumaiya Ali}
\author[1]{Areej Alhothali}
\author[1]{Ohoud Alzamzami}
\author[2]{Sameera Albasri}
\author[3]{Ahmed Abduljabbar}
\author[3]{Muhammad Alwazzan}

\affil[1]{Department of Computer Science, Faculty of Computing and Information Technology,
King Abdulaziz University, Jeddah, Saudi Arabia}

\affil[2]{Faculty of Medicine,
King Abdulaziz University, Jeddah, Saudi Arabia}

\affil[3]{Department of Radiology, King Abdulaziz University Hospital\protect\\Jeddah, Saudi Arabia}

\affil[*]{Corresponding author: sali0174@stu.kau.edu.sa}

\date{}

\begin{document}

\maketitle

\begin{abstract}
Placenta Accreta Spectrum (PAS) is a serious obstetric condition that can be challenging to diagnose with Magnetic Resonance Imaging (MRI) due to variability in radiologists’ interpretations. To overcome this challenge, a hybrid 3D deep learning model for automated PAS detection from volumetric MRI scans is proposed in this study. The model integrates a 3D DenseNet121 to capture local features and a 3D Vision Transformer (ViT) to model global spatial context. It was developed and evaluated on a retrospective dataset of 1,133 MRI volumes. Multiple 3D deep learning architectures were also evaluated for comparison. On an independent test set, the DenseNet121-ViT model achieved the highest performance with a five-run average accuracy of 84.3\%. These results highlight the strength of hybrid CNN-Transformer models as a computer-aided diagnosis tool. The model's performance demonstrates a clear potential to assist radiologists by providing a robust decision support to improve diagnostic consistency across interpretations, and ultimately enhance the accuracy and timeliness of PAS diagnosis.
\end{abstract}
\begin{center}
\end{center}

\section{Introduction}
Placenta Accreta Spectrum (PAS) represents life-threatening obstetric conditions defined by abnormal placental invasion of uterine wall. Its incidence has risen dramatically in recent decades due to rise in cesarean deliveries and the resulting scar tissues \cite{b1, b2}. Estimates indicate that PAS cases have doubled over the last two decades, making PAS a growing health challenge \cite{b3}. The condition carries high risks, including severe hemorrhage, infection, and frequent need for peripartum hysterectomy, contributing significantly to maternal morbidity and mortality \cite{b2}. Early and precise prenatal diagnosis is vital for reducing these risks and allowing multidisciplinary management, which improves outcomes \cite{b1}.

\begin{figure}[htp!]
    \centering
    \includegraphics[width=\textwidth]{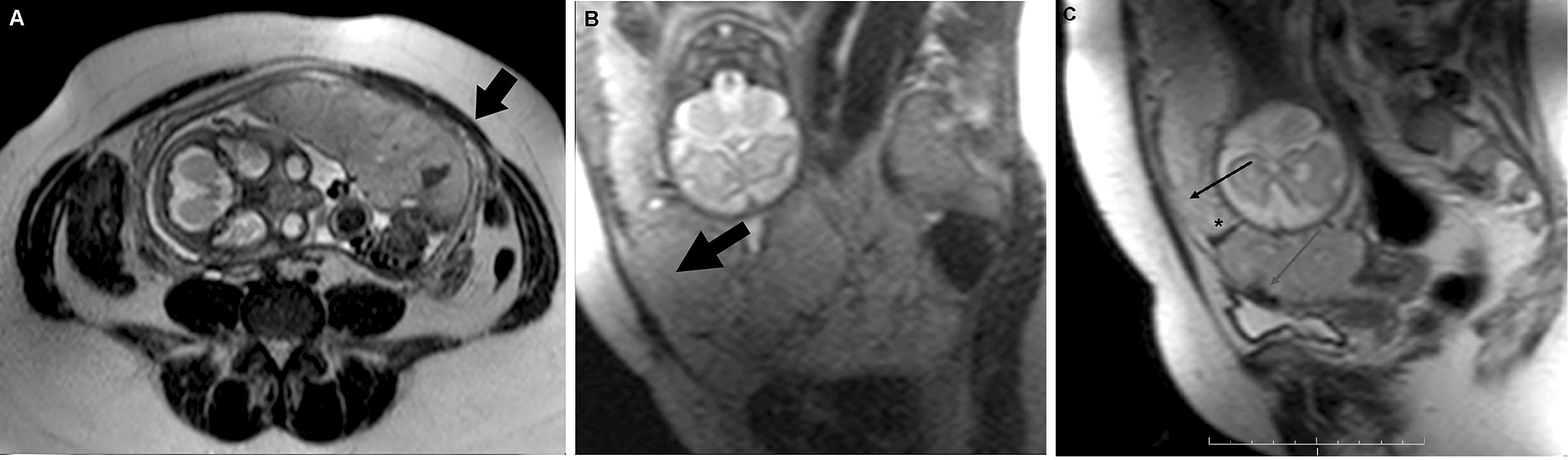}
    \caption{MRI signs of PAS: (A) Intraplacental bands, (B) Myometrial border interruption, and (C) Abnormal vascularity. \cite{b7}}
    \label{fig:MRI_signs}
\end{figure}

Current diagnostic practice combines clinical risk assessment with imaging, primarily ultrasound (US) and magnetic resonance imaging (MRI) \cite{b4}. Although, US imaging is commonly used for initial screening it has limitations in accurately assessing invasion depth and extent, especially with posterior placenta or bowel involvement. MRI offers complementary value through superior soft tissue contrast and larger field of view, providing detailed evaluation of invasion depth and adjacent organ involvement \cite{b2}, \cite{b4}. Among different MRI sequences, T2-weighted imaging (T2WI) is most common for placental assessment and T1-weighted imaging (T1WI) reflect bleeding conditions \cite{b5}. Key MRI signs of PAS include T2-dark intraplacental bands, focal interruption of myometrial border, abnormal vascularity (Fig. \ref{fig:MRI_signs}) \cite{b6, b7}. However, the interpretation of these complex imaging features remain qualitative, requires significant radiological expertise, and prone to inter-observer variability \cite{b8}, presenting the need for objective diagnostic methods.

Deep learning methods like convolutional neural networks (CNNs) have shown notable success in learning complex patterns from raw imaging data, often exceeding human performance in classification and segmentation tasks \cite{b9}. Although CNNs perform well in capturing local features but they often struggle to grasp the global context. Vision Transformers (ViTs) on the other hand uses self-attention mechanisms to effectively extract these global relationships \cite{b10}. This has led to the development of hybrid CNN-Transformer models with the aim to combine the complementary strengths of each approach and show strong performance in complex medical imaging tasks \cite{b11,b12,b13,b14}.

While deep learning has been applied to PAS detection from MRI, systematic studies of 3D architectures remain limited. The optimal model design for capturing both local and global 3D PAS markers is still undefined. This study addresses this gap by proposing a tailored 3D hybrid DenseNet121–ViT model for end-to-end volumetric PAS detection. A systematic comparison of six modern 3D deep learning models was conducted on a novel dataset of 3D MRI. The study demonstrates that integration of dense local feature extraction and global self attention results in superior diagnostic performance.

\section{Related Work}
Computational MRI analysis for PAS has evolved from handcrafted feature engineering to end-to-end deep learning frameworks, with research moving to automated feature extraction \cite{b15}. This reflects broader trends in medical imaging towards automated diagnostic tools. This section reviews prior studies on PAS detection and highlights the performance of hybrid deep learning models in related medical imaging fields.

\subsection{Radiomic Approach}

Radiomics refers to the automated extraction of quantitative imaging features \cite{b15}. Studies show that radiomic features from T2WI MRI can predict PAS with high accuracy. These features including tissue texture, shape, and intensity are extracted using mathematical formulas defined by experts to train machine learning (ML) models. Romeo et al. analyzed MRI-derived texture features from 64 scans (20 PAS) using four ML algorithms, with a k-nearest neighbors classifier achieving 98.1\% accuracy \cite{b16}. Leitch et al. used MRI radiomic features for predicting PAS and hysterectomy risk using 241 scans (141 PAS). After manual uterine and placental segmentation, 17 ML algorithms were tested, achieving 88-92\% accuracy \cite{b17}. A meta-analysis of seven radiomics studies (672 patients) reported pooled sensitivity of 87\%, specificity of 92\%, and area under the receiver operating characteristic curve (AUC) of 0.93, highlighting strong diagnostic potential \cite{b18}. However, feature robustness across different segmentation methods is still a challenge. One study found that using a 3D volume of interest of the retroplacental myometrium resulted the best PAS prediction, confirming the importance of segmentation choice \cite{b19}.

\subsection{Deep Learning}

With advances in deep learning, research has shifted toward automated feature extraction with models like CNN to learn relevant diagnostic patterns directly from raw image data. Early deep learning studies combined radiomic and deep features to improve predictive power. Shao et al. combined radiomic, deep, and clinical features from 112 MRI scans to predict placenta invasion, achieving 94.1\% accuracy and AUC of 0.985. Radiomic features were extracted from manual segmentation and deep features via transfer learning with 3D ResNet50 \cite{b20}. Peng et al. developed a Deep Learning Radiomics (DLR) model for PAS diagnosis using 324 MRI scans (206 PAS), outperforming clinical and radiologist-based models with an AUC of 0.852 on external validation. Radiomic features were extracted with MedicalNet \cite{b21}. Xu et al. proposed a dual-path neural network (parallel ResNet50) for PAS diagnosis combining T1WI and T2WI MRI features via transfer learning on 321 scans (179 PAS), achieving 82.5\% accuracy on held-out test set \cite{b5}.

Recent approaches have employed deep learning for both segmentation and classification using 3D MRI. While Wang et al. achieved a high AUC of 0.897 using a 2D DenseNet-based PAS classifier on 540 cases (170 PAS), their approach relied on inputs from separate 3D segmentation step with 3nnU-Net and a placental position classifier \cite{b22}. Another study used a 3D U-Net 3+ for automatic segmentation of the placenta and uterine cavity on 244 MRI scans, achieving a Dice score of 82.7\% and 91.8\% respectively \cite{b23}. Further research classified PAS subtypes with a two-branched CNN and 414 MRI scans, achieving an AUC of 0.8 for multi-class prediction \cite{b24}. Another study developed a CascadeNet model integrating 3D MRI, radiomic and topographic feature to predict hysterectomy risk, with AUC of 0.878 and 83.3\% accuracy on 241 MRI volumes \cite{b25}. These highlight the potential of fully automated, deep learning systems for PAS assessment.

\subsection{Hybrid Learning}

More recent advances aim to overcome limitations of traditional CNNs through hybrid CNN-ViT models, combining the strengths of both \cite{b10}. These have shown promise in complex medical imaging tasks using MRI, including Alzheimer’s diagnosis \cite{b11,b12} and brain tumor classification \cite{b13, b14}. 

While deep learning has proven effective for MRI-based PAS detection, most studies use small data sample and 2D CNNs. Given the inherent volumetric nature of PAS, this study proposes an end-to-end 3D hybrid CNN-Transformer model that analyzes entire MRI volumes to capture local and global spatial relationships that 2D slice-based or pure 3D CNN analysis may miss. The aim is to assess if a fully volumetric end-to-end approach can perform competitively without needing prior segmentation. Additionally, a comparison with other 3D architectures is performed to validate the superiority of the proposed approach.

\section{MATERIAL AND METHODS}

This study employed a retrospective MRI dataset to develop and evaluate 3D deep learning models for PAS classification. The following subsections describe the dataset and model architectures used.

\subsection{Dataset}

This retrospective study used patient data from the Fetal Medicine Department at King Abdulaziz University Hospital, with all procedures adhering to strict institutional guidelines and ethical standards. Formal ethical approval was obtained from the hospital’s Research Ethics Committee. 

\begin{figure}[htp!]
    \centering
    \includegraphics[width=0.8\textwidth]{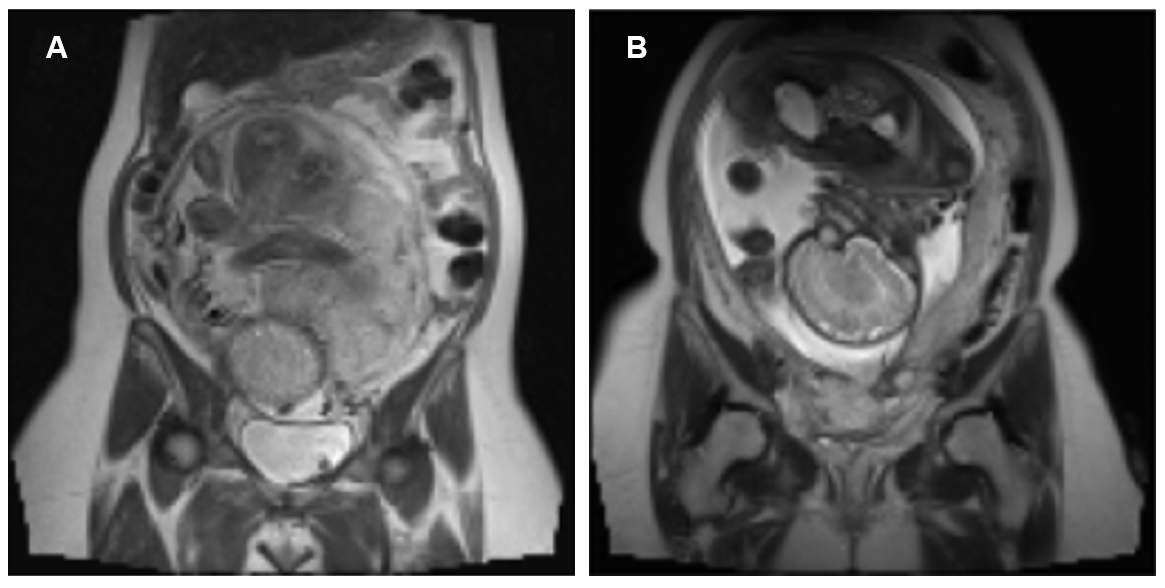}
    \caption{Example MRI slices from the dataset: (A) Normal case, and (B) PAS case.}
    \label{fig:MRI_example}
\end{figure}

An initial query of the hospital’s radiology information system identified a large number of patients who underwent MRI evaluation for suspected PAS. Cases were screened based on diagnostic confirmation and image quality. Exclusion criteria included incomplete imaging, motion-degraded scans, and uncertain diagnostic outcomes. The final dataset consisted of 1,133 T2WI MRI scans, including confirmed 853 normal (non-PAS) and 280 PAS cases, ensuring representation of both diagnostic classes for robust and generalizable deep learning model development. Fig. \ref{fig:MRI_example} illustrates example slices from the dataset.

\subsection{Deep Learning Models}

\begin{figure}[h!]
    \centering
    \includegraphics[width=\textwidth]{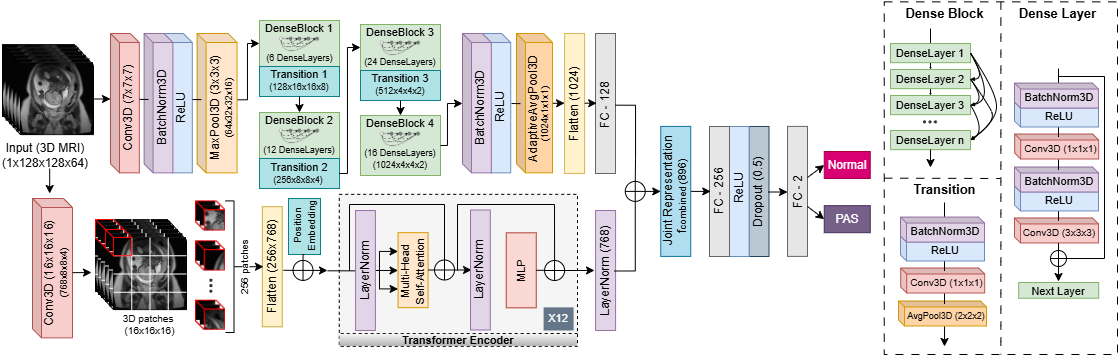}
    \caption{Architecture of the proposed hybrid 3D DenseNet121-ViT model.}
    \label{fig:MRI_architecture}
\end{figure}

In order to extract both local and global contextual features of PAS markers from volumetric MRI, a hybrid 3D DenseNet121-ViT model was developed. A 2D DenseNet-Transformer combination has previously shown promise in medical image classifications \cite{b13}. The proposed hybrid architecture is shown in Fig. \ref{fig:MRI_architecture}. The 3D DenseNet121 component processes the MRI volume to extract local and textural features through a series of dense blocks. A global average pooling layer is applied to the final feature map, followed by a fully connected layer to produce a 128-dimensional embedding. In parallel, the 3D ViT divides the input volume into non-overlapping 16×16×16 patches, which are flattened and linearly embedded. These embeddings pass through 12 transformer blocks with self-attention heads. Layer normalization is applied to the final transformer block output, producing a 768-dimensional embedding. Finally, the two feature representations are concatenated (896-D) and passed to a multilayer perceptron with 50\% dropout for the final binary classification.

A systematic evaluation of multiple 3D deep learning architectures was performed to compare with the proposed architecture. The tested 3D models included: ResNet18, DenseNet121, EfficientNet-B0 and a Swin-Transformer, all known for their effectiveness in medical imaging tasks. The 3D ResNet18 model was pretrained using MedicalNet \cite{b26}. In addition, another hybrid architecture was developed: 3D ResNet18-Swin, selected based on evidence from recent studies \cite{b11, b12}. To reduce overfitting, dropout regularization was applied, with dropout rates between 10\%-50\% evaluated. The optimal dropout rates were found to be 50\% for 3D DenseNet121-ViT and 10\% for 3D ResNet18.

\section{Experimental Setup}

\begin{figure}[htbp]
    \centering
    \includegraphics[width=0.5\textwidth]{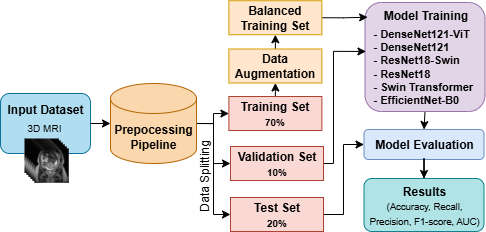}
    \caption{Workflow of the proposed 3D MRI-based PAS classification pipeline.}
    \label{fig:workflow}
\end{figure}

This section provides an overview of the experimental workflow, including data preprocessing, model training, and evaluation. The overall workflow for MRI-based PAS diagnosis is illustrated in Fig. \ref{fig:workflow}.

\subsection{Data Preparation}

\begin{figure}[htbp]
    \centering
    \includegraphics[width=0.8\textwidth]{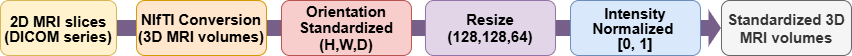}
    \caption{Preprocessing pipeline for standardizing MRI scans.}
    \label{fig:MRI_preprocessing}
\end{figure}

Raw 3D MRI data were initially stored in the Digital Imaging and Communications in Medicine (DICOM) format as a series of 2D slices. Each DICOM series was converted into the Neuroimaging Informatics Technology Initiative (NIfTI) format to create a single 3D volumetric file suitable for 3D CNN input. Volumes were reoriented to a standard (height, width, depth) order to eliminate variability from differing scan orientations. All volumes were resized to a fixed dimension of 128×128×64 voxels using cubic interpolation, with zero-padding to preserve aspect ratio. Finally, voxel intensities were normalized to the range [0, 1] using per scan min-max scaling to reduce scanner-dependent variability while preserving relative intensity patterns within each volume. This standardized preprocessing ensures consistent input sizes for all scans (see Fig. \ref{fig:MRI_preprocessing}).

A noticeable class imbalance was present in the dataset, with 853 normal (denoted as 0) and 280 PAS (denoted as 1) scans. The dataset was split into training (70\%, n=793), validation (10\%, n=113), and independent test sets (20\%, n=227) using stratified sampling to prevent sampling bias and preserve original class distribution across all subsets. it was also ensured that no patient appeared in multiple subsets. The detailed distribution is presented in Table \ref{tab:dataset_summary}.

\begin{table}[htbp]
\centering
\scriptsize
\caption{Dataset distribution across training, validation, and test sets.}
\label{tab:dataset_summary}
\begin{adjustbox}{scale=1.2}
\begin{tabular}{l c c c c}
\hline
\textbf{Class} & \textbf{Training Set} & \textbf{Validation Set} & \textbf{Test Set} & \textbf{Total} \\
\hline
Normal (0) & 597 & 85 & 171 & 853 \\
PAS (1) & 196 & 28 & 56 & 280 \\
\textbf{Total} & \textbf{793} & \textbf{113} & \textbf{227} & \textbf{1133} \\
\hline
\end{tabular}
\end{adjustbox}
\end{table}

The class imbalance in the training set was addressed by augmenting and oversampling the minority class (PAS). The PAS samples were increased from 196 to 597 samples, resulting in a balanced training set of 1,194 volumes. The geometric augmentations applied included random flips along the height and width axes, rotations (90°, 180°, 270°), and zooming (factors of 1.1-1.3).

\subsection{Model Training and Optimization}

All experiments were conducted using the PyTorch and MONAI framework \cite{b27}. Models were trained for 100 epochs and validation performance was monitored after every epoch. The weights achieving the highest validation accuracy were saved for final evaluation on the independent test set using standard binary classification metrics to ensure that the reported performance reflect the model's ability to generalize to unseen data. Adam optimizer was used for all experiments due to its adaptive learning rate. To ensure a fair comparison, hyperparameters were systematically tuned and the final settings used for \ref{tab:hyperparameters}.  

\begin{table}[htbp]
\centering
\scriptsize
\caption{Hyperparameters and their optimized values used in this study.}
\label{tab:hyperparameters}
\setlength{\tabcolsep}{6pt}
\begin{adjustbox}{scale=1.2}
\begin{tabular}{ll}
\hline
\textbf{Hyperparameters}        & \textbf{Value}  \\ \hline
Optimizer       & Adam                        \\
Initial LR      & 0.0001                         \\
LR scheduler    & ReduceLROnPlateau                \\
Loss function   & Cross Entropy      \\
Batch size      & 8                             \\
Epoch           & 100                        \\
\hline
\end{tabular}
\end{adjustbox}
\end{table}

\section{Results and Discussion}

The diagnostic performance of the six evaluated 3D deep learning architectures was assessed on the test set using multiple evaluation metrics: Accuracy, AUC, Precision, Recall, and F1-Score. Each model was trained and tested in five independent runs and the average performance was reported. The results are summarized in Table \ref{tab:model_results} showing a clear performance hierarchy between the models and highlighting the superiority of the proposed hybrid 3D DenseNet121–ViT architecture in all evaluation metrics. 

\begin{table}[htbp]
\centering
\caption{Performance summary of the six evaluated 3D MRI-based models on the independent test set. Best results are highlighted in bold.}
\label{tab:model_results}
\renewcommand{\arraystretch}{1.1}
\setlength{\tabcolsep}{5pt}
\footnotesize
\begin{adjustbox}{width=\textwidth}
\begin{tabular}{lccccc}
\hline
\textbf{Model} & \textbf{Accuracy (\%)} & \textbf{AUC} & \textbf{Precision} & \textbf{Recall} & \textbf{F1-Score} \\
\hline
\pmb{DenseNet121-ViT} & \pmb{85.0 ($84.3 \pm 1.3$)} & \pmb{0.862 ($0.842 \pm 0.012$)} & \pmb{0.799 ($0.790 \pm 0.013$)} & \pmb{0.859 ($0.842 \pm 0.013$)} & \pmb{0.818 ($0.808 \pm 0.014$)} \\
DenseNet121 & 82.8 ($79.5 \pm 2.0$) & 0.804 ($0.766 \pm 0.026$) & 0.770 ($0.732 \pm 0.023$) & 0.802 ($0.764 \pm 0.025$) & 0.783 ($0.743 \pm 0.024$) \\
ResNet18 & 80.2 ($79.3 \pm 1.3$) & 0.829 ($0.783 \pm 0.028$) & 0.759 ($0.738 \pm 0.018$) & 0.832 ($0.781 \pm 0.030$) & 0.772 ($0.750 \pm 0.018$) \\
ResNet18-Swin & 71.3 ($70.0 \pm 1.9$) & 0.644 ($0.600 \pm 0.029$) & 0.629 ($0.599 \pm 0.028$) & 0.642 ($0.604 \pm 0.031$) & 0.634 ($0.601 \pm 0.029$) \\
Swin-Transformer & 72.7 ($69.0 \pm 2.8$) & 0.592 ($0.548 \pm 0.029$) & 0.608 ($0.552 \pm 0.034$) & 0.585 ($0.546 \pm 0.028$) & 0.591 ($0.548 \pm 0.030$) \\
EfficientNet-B0 & 64.8 ($62.8 \pm 1.7$) & 0.692 ($0.604 \pm 0.047$) & 0.596 ($0.573 \pm 0.018$) & 0.622 ($0.592 \pm 0.023$) & 0.593 ($0.569 \pm 0.019$) \\
\hline
\multicolumn{6}{l}{\textit{Note:} Values are reported as \textit{Best (Mean ± Standard Deviation)} across five independent runs.}
\end{tabular}
\end{adjustbox}
\end{table}

\begin{figure}[htp!]
    \centering
    \includegraphics[width=0.78\textwidth]{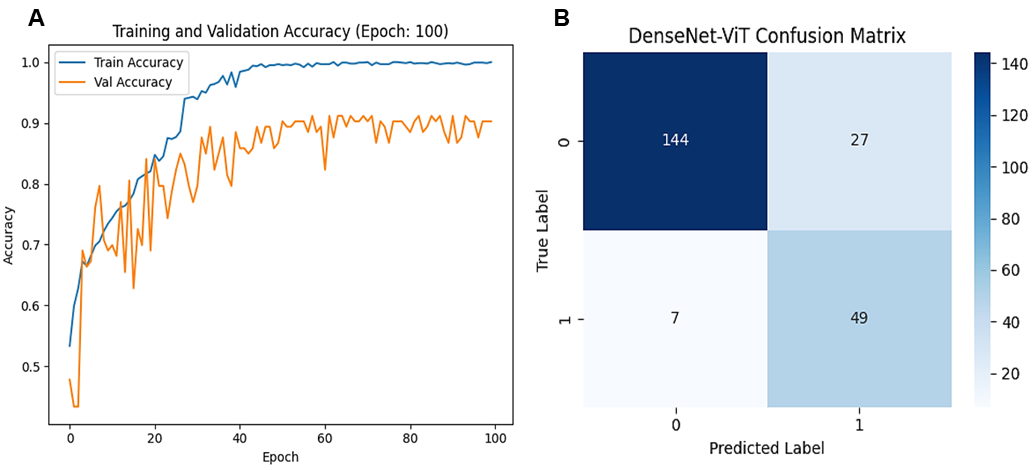}
    \caption{Performance of the best DenseNet121–ViT model: (A) Training and validation curves, and (B) Confusion matrix on the test set.}
    \label{fig:performance}
\end{figure}

The hybrid 3D DenseNet121-ViT model yielded the best overall performance. It achieved a five-run average accuracy of 84.3±1.3\% and an average AUC of 0.842±0.012. This model also demonstrated balanced performance between precision (0.790±0.013) and recall (0.842±0.013), resulting in the highest F1-Score of 0.808±0.014. The low standard deviation (SD) between runs indicates strong stability and robustness, a critical attribute for potential clinical translation. For its best-performing run, the model reached 98.6\% training and 91.2\% validation accuracy at peak validation performance over 100 epochs (Fig. \ref{fig:performance}A). The corresponding test accuracy was 85.0\% with an AUC of 0.862. The confusion matrix indicated that it correctly identified 144 out of 171 normal cases (0) and 49 out of 56 PAS cases (1) in the test set (Fig. \ref{fig:performance}B). 
 
\begin{figure}[htp!]
    \centering
    \includegraphics[width=0.53\textwidth]{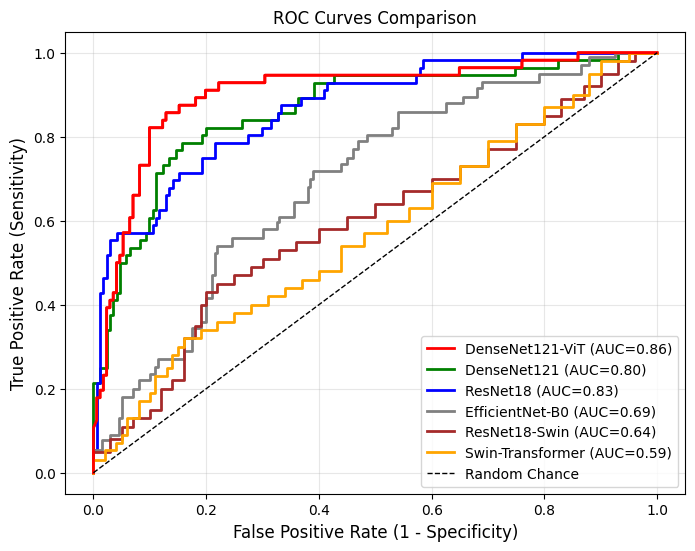}
    \caption{Receiver Operating Characteristic (ROC) curves of six evaluated MRI models.}
    \label{fig:roc_acc}
\end{figure}

The second best performance was of pure 3D DenseNet121 (accuracy 79.5±2.0) and closely matched by 3D ResNet18 (accuracy 79.3±1.3). The remaining architectures resulted in comparatively lower performance, with accuracies ranging from 63\% to 70\%. Fig. \ref{fig:roc_acc} presents the receiver operating characteristic (ROC) curve corresponding to the best-performing runs of each MRI-based model.   

To assess the statistical significance of performance differences among the six models, repeated measures analysis of variance (ANOVA) was conducted, followed by post-hoc paired t-tests. The resulting p-values were corrected for multiple comparisons using the Benjamini–Hochberg procedure to control the false discovery rate (FDR). The analysis confirmed that the DenseNet121-ViT model outperformed the other architectures across all metrics $(p < 0.05)$. Similarly, DenseNet121 and ResNet18 performed significantly better than EfficientNet-B0, ResNet18-Swin, and Swin-Transformer. No significant differences were observed between DenseNet121 and ResNet18, or between ResNet18-Swin and Swin-Transformer $(p > 0.05)$. Significant pairwise differences based on accuracy are summarized in Table \ref{tab:pairwise_ttest}. 

\begin{table}[htbp]
\centering
\caption{Pairwise \textit{t}-test results (accuracy) between the models (\(\alpha = 0.05\)). Each cell shows the FDR corrected \textit{p}-value and significance (\checkmark = significant, $\times$ = not significant).}
\label{tab:pairwise_ttest}
\renewcommand{\arraystretch}{1.1}
\setlength{\tabcolsep}{2pt}
\footnotesize
\begin{center}
\begin{adjustbox}{scale=0.9}
\begin{tabular}{lcccccc}
\hline
\textbf{Model} & \textbf{D121} & \textbf{R18} & \textbf{R18-Swin} & \textbf{Swin} & \textbf{Eff-B0} \\
\hline
\textbf{D121-ViT} & 0.018 (\checkmark) & 0.013 (\checkmark) & $<0.001$ (\checkmark) & 0.002 (\checkmark) & $<0.001$ (\checkmark) \\
\hline
\textbf{D121}  & -- & 0.886 ($\times$) &$<0.001$ (\checkmark) & 0.001 (\checkmark) & $<0.001$ (\checkmark) \\
\hline
\textbf{R18}    & -- & -- & $<0.001$ (\checkmark) & 0.005 (\checkmark) & $<0.001$ (\checkmark) \\
\hline
\textbf{R18-Swin}  & -- & -- & -- & 0.526 ($\times$) & \multicolumn{1}{r}{0.009 (\checkmark)} \\
\hline
\textbf{Swin}  & -- & -- & -- & -- & \multicolumn{1}{r}{0.023 (\checkmark)} \\
\hline 
\end{tabular}
\end{adjustbox}
\end{center}
\vspace{-8pt}
\raggedright
\scriptsize
\textit{Abbreviations:} D121-ViT: DenseNet121-ViT; D121: DenseNet121; R18: ResNet18; R18-Swin: ResNet18-Swin; Swin: Swin-Transformer; Eff-B0: EfficientNet-B0.
\end{table}

The proposed model’s success can be attributed to its ability to harness the complementary strengths of its two components. The DenseNet121 component, through dense connectivity and feature reuse, may be effective at identifying local fine-grained textures such as T2-dark intraplacental bands (Fig. 1A). Meanwhile, the ViTs global self-attention is well-suited for modeling long-range spatial relationships, such as the overall shape and integrity of the myometrial border across the uterine volume (Fig. 1B). In essence, the dual-scale processing imitates an expert radiologist’s approach, explaining its superior performance compared to the single-scale CNN or Transformer models. Notably, the performance of the ResNet18-Swin hybrid, which was worse than the standalone ResNet18, suggests that simply combining a CNN and a Transformer is not a guaranteed formula for success. The effectiveness of a hybrid model is critically dependent on the specific pairing and fusion strategy. This failure may be due to a mismatch between ResNet18 and Swin-Transformer features or the combined model's parameter count was too large causing overfitting on the dataset. This highlights that the DenseNet121-ViT combination proved particularly effective for this task.

Compared with prior MRI-based PAS studies, the proposed model’s results are competitive (average AUC 0.842, best-run AUC 0.862). Other studies for PAS diagnosis with deep learning have reported external validation AUCs from 0.849 to 0.897. However, these often relied on handcrafted features, 2D analysis, or limited datasets $(< 550)$. This study extends the literature by demonstrating the effectiveness of an end-to-end volumetric hybrid model developed on a relatively larger dataset of $> 1,100$ MRI volume to allow robust learning and generalization while avoiding biases from manual feature engineering. Thus, the fully automated design and 3D approach represent a more clinically scalable solution.

\section{Conclusion}

This study presented a 3D DenseNet121-ViT model to effectively classify PAS on MRI data and achieve robust accuracy and AUC. The proposed model offers a key advantage over 2D slice-based methods by performing an end-to-end volumetric analysis without losing important global context. The combination of local feature extraction and global context allows the hybrid DenseNet121-ViT framework to outperform standalone 3D CNNs. This represents an important step in improving automated prenatal image analysis. The proposed model as a computer-aided diagnosis instrument demonstrates a strong performance, indicating its clinical relevance. Clinically it can assist radiologists by providing objective risk assessments, improve diagnostic consistency while reducing the observer variability that currently challenges PAS diagnosis.  In short, such a tool would be clinically deployed through an integration with the hospital's radiology information system software. The model may be used in this workflow to analyze MRI volumes automatically for screening prenatal high-risk patients with PAS. The radiologist would then be alerted with a positive or high-risk classification to act as an effective diagnostic aid to support the final diagnosis and the multidisciplinary surgical planning.

Some limitations should be noted in this study. The retrospective, single institution dataset may introduce selection bias and limit generalizability across scanners and populations. Future research should focus on multi-center validation and integration of multimodal data. Finally, the ”black box” nature of the model could be a barrier to clinical adoption. Therefore,the future work will involve the creation of interpretability methods, including visualizations that show which parts of the image most affected the model's decision, which is essential in creating clinical trust in such an automated system. By addressing these challenges, hybrid deep learning approaches hold promise for advancing non-invasive, accurate prenatal PAS diagnosis and reliable automated clinical decision support.

\section*{Acknowledgments}
The dataset collected for this study was approved by the Research Ethics Committee at King Abdulaziz University (Reference No. HA-02-J-008) on September 11, 2023, ensuring compliance with ethical guidelines for participant rights and confidentiality.


\begin{thebibliography}{27}

\bibitem{b1} Einerson, B.D., Gilner, J.B., Zuckerwise, L.C.: Placenta accreta spectrum. Obstet. Gynecol. 142(1), 31–50 (2023). doi:10.1097/AOG.0000000000005229

\bibitem{b2} Cahill, A.G., Beigi, R., Heine, R.P., Silver, R.M., Wax, J.R.: Placenta accreta spectrum. Am. J. Obstet. Gynecol. 219(6), B2–B16 (2018). doi:10.1016/j.ajog.2018.09.042

\bibitem{b3} Jauniaux, E., Bunce, C., Grønbeck, L., Langhoff-Roos, J.: Prevalence and main outcomes of placenta accreta spectrum: a systematic review and meta-analysis. Am. J. Obstet. Gynecol. 221(3), 208–218 (2019). doi:10.1016/j.ajog.2019.01.233

\bibitem{b4} Arakaza, A., Zou, L., Zhu, J.: Placenta accreta spectrum diagnosis challenges and controversies in current obstetrics: A review. Int. J. Women’s Health 15, 635–654 (2023). doi:10.2147/IJWH.S395271

\bibitem{b5} Xu, J., Shao, Q., Chen, R., Xuan, R., Mei, H., Wang, Y.: A dual-path neural network fusing dual-sequence magnetic resonance image features for detection of placenta accreta spectrum (PAS) disorder. Math. Biosci. Eng. 19(6), 5564–5575 (2022). doi:10.3934/mbe.2022260

\bibitem{b6} Concatto, N.H., Westphalen, S.S., Vanceta, R., Schuch, A., Luersen, G.F., Ghezzi, C.L.A.: Magnetic resonance imaging findings in placenta accreta spectrum disorders: pictorial essay. Radiol. Bras. 55(3), 181–187 (2022). doi:10.1590/0100-3984.2021.0115

\bibitem{b7} Romeo, V. et al.: Prediction of placenta accreta spectrum in patients with placenta previa using clinical risk factors, ultrasound and magnetic resonance imaging findings. La Radiol. Med. 126(9), 1216–1225 (2021). doi:10.1007/s11547-021-01348-6

\bibitem{b8} Parnes, B. et al.: Reproducibility of MRI features of placenta accreta spectrum disorders. Radiology 311(2), e240386 (2024). doi:10.1148/radiol.240386

\bibitem{b9} Mienye, I.D., Swart, T.G., Obaido, G., Jordan, M., Ilono, P.: Deep convolutional neural networks in medical image analysis: A review. Information 16(3), 195 (2025). doi:10.3390/info16030195

\bibitem{b10} Takahashi, S. et al.: Comparison of vision transformers and convolutional neural networks in medical image analysis: A systematic review. J. Med. Syst. 48, 84 (2024). doi:10.1007/s10916-024-02105-8

\bibitem{b11} Zhao, Z. et al.: Vision transformer-equipped convolutional neural networks for automated Alzheimer’s disease diagnosis using 3D MRI scans. Front. Neurol. 15, 1490829 (2024). doi:10.3389/fneur.2024.1490829

\bibitem{b12} Zhou, J. et al.: A deep learning model for early diagnosis of Alzheimer’s disease combined with 3D CNN and video Swin transformer. Sci. Rep. 15, 23311 (2025). doi:10.1038/s41598-025-05568-y

\bibitem{b13} Aloraini, M., Khan, A., Aladhadh, S., Habib, S., Alsharekh, M.F., Islam, M.: Combining the transformer and convolution for effective brain tumor classification using MRI images. Appl. Sci. 13(6), 3680 (2023). doi:10.3390/app13063680

\bibitem{b14} Shanto, M.N.I., Mubtasim, M.T., Rakshit, S.V., Ullah, M.A.: Enhanced classification of brain tumors from MRI scans using a hybrid CNN-transformer model. In: Proc. 2025 Int. Conf. Quantum Photonics, Artif. Intell., Netw. (QPAIN), pp. 1–6. Rangpur, Bangladesh (2025). doi:10.1109/QPAIN66474.2025.11171896

\bibitem{b15} Danaei, M. et al.: Machine learning applications in placenta accreta spectrum disorders. Eur. J. Obstet. Gynecol. Reprod. Biol. X 25, 100362 (2024). doi:10.1016/j.eurox.2024.100362

\bibitem{b16} Romeo, V. et al.: Machine learning analysis of MRI-derived texture features to predict placenta accreta spectrum in patients with placenta previa. Magn. Reson. Imaging 64, 71–76 (2019). doi:10.1016/j.mri.2019.05.017

\bibitem{b17} Leitch, K. et al.: Placenta accreta spectrum and hysterectomy prediction using MRI radiomic features. In: Proc. SPIE Int. Soc. Opt. Eng. 12033, 120331I (2022). doi:10.1117/12.2611587

\bibitem{b18} Huang, L. et al.: Accuracy of MRI-based radiomics in diagnosis of placenta accreta spectrum: A PRISMA systematic review and meta-analysis. Med. Sci. Monit. 30, e943461 (2024). doi:10.12659/MSM.943461

\bibitem{b19} Verde, F. et al.: Segmentation methods applied to MRI-derived radiomic analysis for the prediction of placenta accreta spectrum in patients with placenta previa. Abdom. Radiol. 48, 3207–3215 (2023). doi:10.1007/s00261-023-03963-5

\bibitem{b20} Shao, Q. et al.: Deep learning and radiomics analysis for prediction of placenta invasion based on T2WI. Math. Biosci. Eng. 18(5), 6198–6215 (2021). doi:10.3934/mbe.2021310 

\bibitem{b21} Peng, L. et al.: Prenatal diagnosis of placenta accreta spectrum disorders: Deep learning radiomics of pelvic MRI. J. Magn. Reson. Imaging 58(5), 1541–1553 (2023). doi:10.1002/jmri.28787

\bibitem{b22} Wang, H. et al.: A deep learning pipeline using prior knowledge for automatic evaluation of placenta accreta spectrum disorders with MRI. J. Magn. Reson. Imaging 58(4), 1221–1230 (2023). doi:10.1002/jmri.28770

\bibitem{b23} Huang, J. et al.: Deep learning based automatic segmentation of the placenta and uterine cavity on prenatal MR images. In: Proc. SPIE 12465, Medical Imaging 2023: Image Processing, 124650N (2023). doi:10.1117/12.2653659

\bibitem{b24} Jiang, H., Liu, Q., Zhou, Y., Pan, J., Song, T., Lu, Y.: Anatomy-guided multitask learning for MRI-based classification of placenta accreta spectrum and its subtypes. In: Proc. IEEE 22nd Int. Symp. Biomed. Imaging (ISBI), pp. 1–5 (2025). doi:10.1109/ISBI60581.2025.10981099

\bibitem{b25} Dormer, J.D. et al.: CascadeNet for hysterectomy prediction in pregnant women due to placenta accreta spectrum. In: Proc. SPIE 12032, Medical Imaging 2022: Image Processing, 120320N (2022). doi:10.1117/12.2611580 

\bibitem{b26} Chen, S., Ma, K., Zheng, Y.: Med3D: Transfer learning for 3D medical image analysis. arXiv preprint (2019). doi:10.48550/arXiv.1904.00625

\bibitem{b27} Cardoso, M.J. et al.: MONAI: An open-source framework for deep learning in healthcare. arXiv preprint (2022). doi:10.48550/arXiv.2211.02701

\end{thebibliography}
\end{document}